%% file: template.tex
\documentclass[electronic]{vgtc}             

\ifpdf
  \pdfoutput=1\relax                   
  \usepackage{graphicx}                
  \DeclareGraphicsExtensions{.pdf,.png,.jpg,.jpeg} 
\else
  \usepackage{graphicx}                
  \DeclareGraphicsExtensions{.eps}     
\fi%

\graphicspath{{figures/}{pictures/}{images/}{./}} 

\usepackage{microtype}                 
\PassOptionsToPackage{warn}{textcomp}  
\usepackage{textcomp}                  
\usepackage{mathptmx}                  
\usepackage{times}                     
\usepackage{cite}                      
\usepackage{tabu}                      
\usepackage{booktabs}                  
\usepackage{flushend}
\usepackage[symbol]{footmisc}

\onlineid{1175}

\vgtccategory{Research}

\vgtcinsertpkg

\title{Immersive Neural Graphics Primitives}

\newcommand\Mark[1]{\textsuperscript#1}

\author{ \textbf{Ke Li} \Mark{{1, 3}} $^*$ %
\and \textbf{Tim Rolff} \Mark{{1, 2}} \thanks{These authors contributed equally to the work.}%
\and Susanne Schmidt \Mark{1}%
\and Reinhard Bacher  \Mark{3} %
\and Simone Frintrop \Mark{2} %
\and Wim Leemans \Mark{3}%
\and Frank Steinicke \Mark{1}}
\affiliation{
\scriptsize 
\Mark{1} Human-Computer Interaction Group, Department of Informatics, Universität Hamburg \\
\Mark{2} Computer Vision Group, Department of Informatics, Universität Hamburg \\
\Mark{3} Deutsches Elektronen-Synchrotron DESY, Germany \\} 

\teaser{
  \centering
  \includegraphics[width=0.8\linewidth]{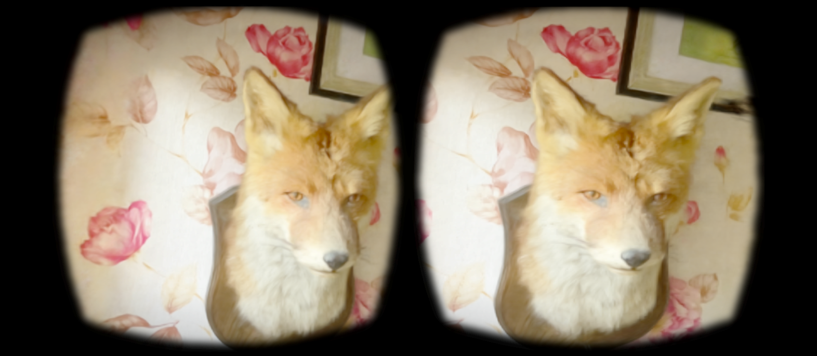}
  \caption{Illustration of the standard evaluation fox scene from the instant neural graphics primitives (instant-ngp) framework with DLSS support rendered for a head-mounted display using the Unity engine.}
  \label{fig:teaser}
}

\abstract{
Neural radiance field (NeRF), in particular its extension by instant neural graphics primitives, is a novel rendering method for view synthesis that uses real-world images to build photo-realistic immersive virtual scenes. 
Despite its potential, research on the combination of NeRF and virtual reality (VR) remains sparse. 
Currently, there is no integration into typical VR systems available, and the performance and suitability of NeRF implementations for VR have not been evaluated, for instance, for different scene complexities or screen resolutions. 

In this paper, we present and evaluate a NeRF-based framework that is capable of rendering scenes in immersive VR allowing users to freely move their heads to explore complex real-world scenes.
We evaluate our framework by benchmarking three different NeRF scenes concerning their rendering performance at different scene complexities and resolutions.
Utilizing super-resolution, our approach can yield a frame rate of 30 frames per second with a resolution of 1280$\times$720 pixels per eye.
We discuss potential applications of our framework and provide an open source implementation online.\footnote{Link to the repository: https://github.com/uhhhci/immersive-ngp}

} 

\CCScatlist{
  \CCScatTwelve{Neural Radiance Field}{Instant Neural Graphics Primitives}{Immersive Experiences}{Free-viewpoint Videos};
}

\begin{document}


\firstsection{Introduction}

\maketitle

Advancements in head-mounted displays (HMDs) and graphics processing units (GPUs) allow for immersive virtual reality (VR) experiences, which can realistically display objects or scenes from the real world.
Such accurate 3D representations of real-world environments are crucial for a variety of immersive VR applications. For example, immersive journalism \cite{realistic_immersive_journalism, Immersive_journalism2}, robot teleoperation \cite{MR-robot-teleoperation}, cultural participation \cite{schmidt2019effects}, and entertainment \cite{Realism_gaming} could benefit from having an accurate photo-realistic 3D representation of the real world in which users can freely move instead of a 360 degree video with a fixed viewpoint. 
However, after acquiring the images or videos of real environments, state-of-the-art 3D reconstruction pipelines require computationally expensive processes based on established techniques such as photogrammetry \cite{photogrammetry} or simultaneous localization and mapping \cite{SLAM}, limiting their applicability, especially for large scenes \cite{swarmSLAM}.
Other methods for 3D scene reconstruction rely on expensive hardware, such as light detection and ranging  \cite{LIDAR} or structured light sensors \cite{structured-light-3D-scanning}.
Moreover, manual post-processing is required in multiple of these techniques. 
As a result, the conventional process for high-quality 3D scene reconstruction is time-consuming and requires expert knowledge in the field of 3D modeling.

Recently, \emph{neural radiance field (NeRF)} \cite{NERForiginal, instant-ngp} has rapidly emerged as a method, which enables high-quality reconstruction of photo-realistic 3D scenes based on multiple 2D images and their camera poses only. 
While the original work by Mildenhall et al. \cite{NERForiginal} requires extensive computation for training a single scene, Müller et al. \cite{instant-ngp}  recently proposed \emph{instant neural graphics primitives \mbox{(instant-ngp)}} with a multi-resolution hash encoding, which simultaneously enables rapid training and real-time, high-resolution neural radiance rendering.

\begin{figure*}
    \centering
    \includegraphics[width =0.98\linewidth]{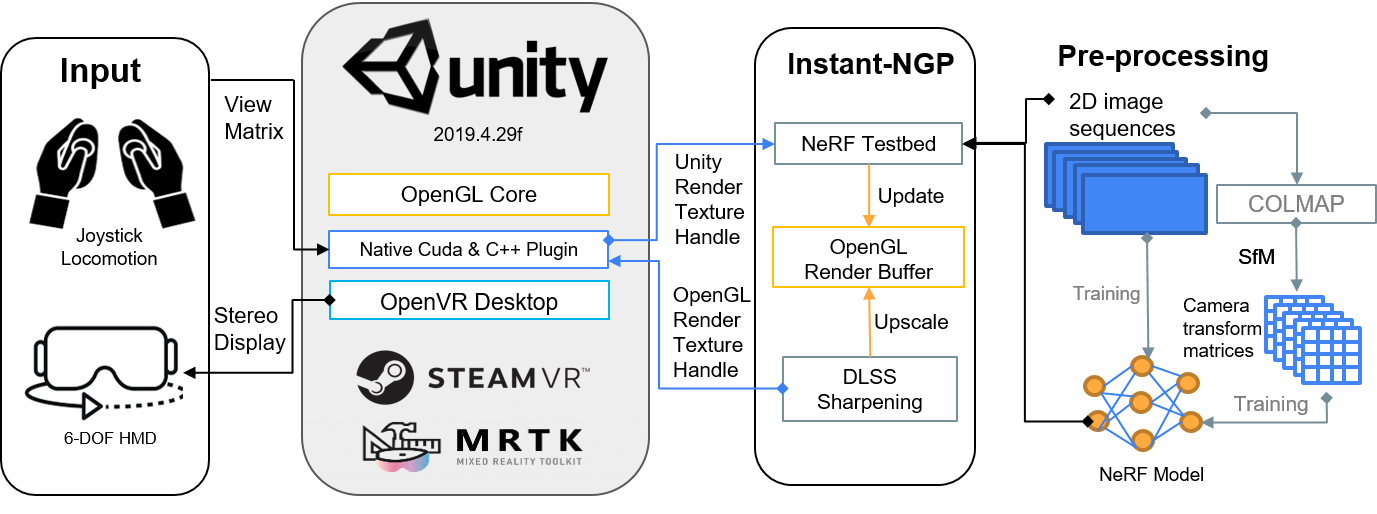}
    \caption{The system architecture of our framework, depicting the individual processes from left to right: Starting with the input, a view matrix is computed from VR input devices and the HMD. This view matrix is then applied in the native Unity plugin, which provides a communication layer to the instant-ngp backend as well as the final rendering. The instant-ngp backend performs the volume rendering through NeRF by updating the provided texture. The pre-processing refers to the model training.}
    \label{fig:system_setup}
\end{figure*}

Although, NeRF allows for efficient, image-based 3D scene reconstruction, which provides enormous potential for various VR applications, unfortunately, there is currently no open-source NeRF framework available for the VR research community which is integrated into existing VR rendering engines, such as Unity.
In fact, current NeRF implementations are developed mostly for proof-of-concept computer vision and machine learning research.\\

To bridge this current gap in the availability of NeRF software systems, we develop \emph{immersive instant-ngp}, an adaptation of the original instant-ngp framework \cite{instant-ngp} with an integration into Unity, which is one of the most widely used game engines for VR research and development. Our system simultaneously renders two instances of high-resolution NeRF images in Unity with the support of Deep Learning Super Sampling (DLSS) \cite{DLSS}, creating a stereoscopic video see-through (VST) of the NeRF scene in VR. 
In addition, \emph{immersive instant-ngp} enables free-viewpoint exploration, manual adjustment of the bounding box, and six degrees of freedom (DoF) manipulation of a scene. 
To our best knowledge, this is the first open-source NeRF framework for VR-based on Unity.\\  

\noindent The contributions of our work can be summarized as follows:
\begin{itemize}
    \item We provide the first open-source framework for rendering immersive photo-realistic neural graphics primitive scenes in VR supporting frame rates around 30 frames per second (fps).  
    \item We evaluate our proposed framework for different scene complexities and resolutions with regard to frame timings and frame rates.
\end{itemize}
The remainder of this paper is structured as follows.
Section \ref{sec:Related Work} resumes work related to NeRF and free-viewpoint videos. 
Section \ref{sec:Software Implementation} describes the architecture and available features of our system and framework. 
Section \ref{sec:Performance Benchmarking} shows the results of a benchmark performance of our immersive instant-ngp system. 
Section \ref{sec:Limitations} discusses the systems and its limitations. 
Section \ref{sec:Conclusion} concludes our work and gives an overview about future work of NeRF for VR.

\section{Related Work}
\label{sec:Related Work}
\label{sec:Software Implementation}
Free-viewpoint videos (FVVs) are an important extension of 360 degree videos for immersive VR, because viewers can freely navigate through the scene and change their view orientation as well as their position within the scene \cite{FVV-Microsoft}.
Most of today's FVVs for VR are based on cameras which capture the entire 360 degree field of view (FoV) from a single point in the center \cite{omnidirectional_camera}. 
Therefore, 360 degree videos do only support three DoF viewing experiences, i.e., changes of the orientation, whereas changes of position are not supported.
Other methods for creating six DoF FVV videos typically require a computationally expensive offline pre-processing stage for dense 3D scene reconstruction \cite{FVV-Microsoft, 6DOF-360-videos}. 
To synthesize novel views from only sparse image information supporting full six DoF FVV remains a major research challenge in the VR, computer vision, and computer graphics communities. 

Recently, a breakthrough in NeRF research opens up new possibilities for view synthesis and efficient FVV creation \cite{NERForiginal}.  
NeRF utilizes a continuous volumetric scene function to represent the underlying scene through a neural network, taking the spatial position and viewing direction as inputs.
In contrast to conventional 3D measurement techniques, which require point-cloud calculation, NeRF learns a volumetric representation \cite{VolumetricRendering} to predict the color of every pixel by learning the volumetric density and view-dependent emitted radiance at the volumetric spatial input location \cite{NERForiginal}. 
Rendering results can be further improved by encoding the spatial input position into a high-frequency signal. 
This position encoding enables the neural network to learn high-frequency functions in low-dimensional domains \cite{FourierFeatureNetwork} such that high-resolution, photo-realistic 3D scene reconstruction gets possible. 

The rapid development of various view synthesis techniques based on NeRF further enables extended efficient volumetric 3D scene reconstruction options, such as volumetric FVV for dynamic scenes \cite{d-NeRF}, FVV for large-scale scenes \cite{block-NeRF}, and volumetric scene reconstruction based on unconstrained photographs \cite{NeRF-in-the-wild}. 
Furthermore, instant-ngp, an extension of the NeRF research, already enables a real-time application of NeRF \cite{instant-ngp}. It introduces a new hash function to encode the spatial position, 
enabling training of high-quality neural graphics primitives in seconds. 

While NeRF promises enormous potentials for VR applications, there is still very limited research in using NeRF in the VR community to date. 
An exception is the work by Deng et al., who recently proposed a foveated rendering approach for NeRF \cite{FoV-NeRF}.
However, currently there is no VR implementation for instant-ngp. Moreover, there are no open-source NeRF frameworks available for the integration into VR systems.
Finally, previous work has not evaluated the NeRF method for different scene complexities or resolutions when applied in VR, which makes it hard for VR researchers to assess whether the NeRF method provides a reasonable alternative for the traditional rendering pipeline.

To bridge this gap, we introduced the first open-source VR NeRF framework (\emph{immersive instant-ngp}), which enables efficient, high-quality, and stereoscopic FVV experiences in VR.

\section{Immersive instant-ngp Framework}
\label{sec:Software Implementation}

\subsection{System Architecture}

\paragraph{Framework Overview:} Figure \ref{fig:system_setup} gives an overview of the architecture of our framework. 
It renders a NeRF 3D scene based on a pre-trained instant-ngp model with compressed volumetric scene representation. 
The Unity application uses input from the six DoF tracking of the VR HMD as well as its controllers.
Based on these transformations, a corresponding view matrix describing the camera rotation and translation is applied to the instant-ngp instance and the scene is rendered accordingly. 
Finally, the Unity OpenVR desktop plugin displays the rendered content on the VR HMD.   

\paragraph{Image Pre-processing:} To generate a neural radiance representation of a 3D scene from a set of unordered 2D images, camera poses need to be estimated. The required transformation matrices are typically computed using conventional photogrammetry tools such as COLMAP \cite{COLMAP}, which integrates a general purpose structure from motion (SfM) algorithm \cite{SFM-revisited} that can iteratively reconstruct 3D camera transforms based on common features extraction. As NeRF depends on accurate camera pose estimation for precise positional encoding of the scene. Therefore, having a reasonable scene overlap from the input images set is important for the quality of the NeRF scene reconstruction and rendering \cite{instant-ngp}. The recommended input image size for instant-ngp is between 50-150 images with sparse views focusing on a reconstruction target or area \cite{instant-ngp}. 

\paragraph{The Unity Application:} The main VR application is driven by Unity with the OpenVR Desktop runtime \cite{OpenVR}. Currently, the original instant-ngp framework \cite{instant-ngp} creates render context using OpenGL to provide support for both Linux and Windows platforms. Although OpenXR proposed by the Khronos Group is becoming the de facto standard for cross-platform VR~\cite{OpenXR}, to date, the OpenXR runtime of many popular VR headsets such as the Oculus Quest 2 does not support OpenGL in Unity Windows platforms. To adapt for instant-ngp's dependency on OpenGL, the runtime we use to access different VR hardware is a fallback to OpenVR running together with Steam VR\footnote{https://store.steampowered.com/app/250820/SteamVR/}. Therefore, our framework currently runs on Unity version 2019.4.29f1, which is compatible with both the OpenVR desktop and the OpenGL graphics API. 

\paragraph{Instant-ngp Native Unity Plugin:} 
To support direct communication between Unity and the instant-ngp application, we developed a native Unity C++ plugin, which provides a Unity runtime with interfaces to instant-ngp through pre-compiled dynamic link libraries (DLL). 
As both the Unity Windows VR application and instant-ngp runs on OpenGL, a single render buffer per eye is instantiated at the beginning of the runtime and shared between the Unity application and an instant-ngp instance. To update the render texture in Unity, only native pointer handles need to be shared to the instant-ngp render buffer through a native C++ function call. Render texture handles can likewise be efficiently sent back to Unity to trigger event callbacks. 
Moreover, devices with Nvidia DLSS support can enable super-resolution and sampling, and the render buffer can be directly up-scaled from the Unity runtime.

\paragraph{Stereo Rendering:} 
The render buffer simultaneously renders two Unity textures of cameras that are positioned with a horizontal separation adjustable to any interpupillary distance (IPD) value to accommodate for different distances between users' eyes. 
This supports stereoscopic perception of the scene with an immersive experience.
Rendering resolution and the target frame rate can be freely adjusted according to the performance capability of different GPU hardware.

\subsection{Six DoF VR Interactions}

\paragraph{Six DoF Exploration} To support smooth, uninterrupted movement in the VE, our framework implements continuous locomotion, where users could either travel by natural walking, or by controller-based navigation for stationary usage \cite{VRLocomotion}. While controller-based continuous locomotion can maintain a simple way of moving through the VE, it can also lead to motion sickness due to sensory conflicts \cite{teleportationVR}. 
To reduce these potential conflicts, many immersive VR games implement a simple "point and teleport" technique \cite{teleportationVR}. 
However, as NeRF only creates a volumetric representation of the scene, there is a lack of geometrical collision data for way-point selection, thus, the support of conventional VR teleportation requires additional measures. 

\paragraph{Six DoF Manipulation}

\begin{figure}
    \centering
    \includegraphics[width  =\linewidth]{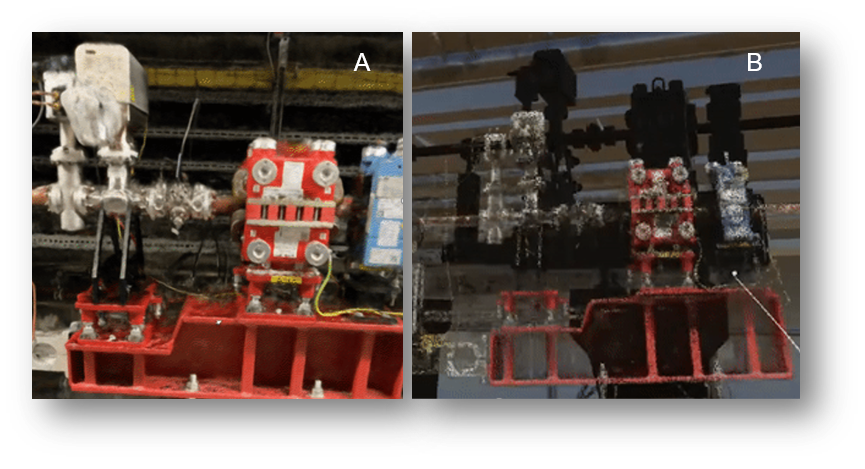}    \caption{Example of using offline volume image slices to augment an existing abstract geometric representation of a scene, such as a 3D CAD model.}
    \label{fig:volume_data_manipulation}
\end{figure}

In addition to the egocentric online neural graphics rendering capability, our framework also supports 3D manipulation of volume image slices. These slices are the density values along a two-dimensional axis-aligned plane of the original volume. As the instant-ngp framework could directly generate such volume data, we could directly export the image slices and import them to the Unity application. Our Unity framework supports direct volume ray-casting algorithm \cite{directVolumeRaycast}, adapted from the Unity Volume Rendering package \cite{UnityVolumeRendering}. Users can interact with and manipulate the volume data via controllers to re-scale, rotate, and re-position it. 
As illustrated in Figure \ref{fig:volume_data_manipulation}B, a miniature representation of the scene and manipulation of the 3D scene via an exocentric perspective provides users with an overview of the scene when experiencing a large-scale NeRF environment. 
Furthermore, \mbox{Figure \ref{fig:volume_data_manipulation}} illustrates how the offline volume image slices can be used to augment an existing abstract geometrical 3D structure, such as an industrial CAD model. 
The volume data generated by instant-ngp can be manipulated such that they are completely overlaid with the 3D geometries, creating a digital twin that has both realistic textures and geometries.

\section{Performance Benchmarking}
\label{sec:Performance Benchmarking}
\begin{figure*}[!ht]
    \centering
    \includegraphics[width=0.97\linewidth]{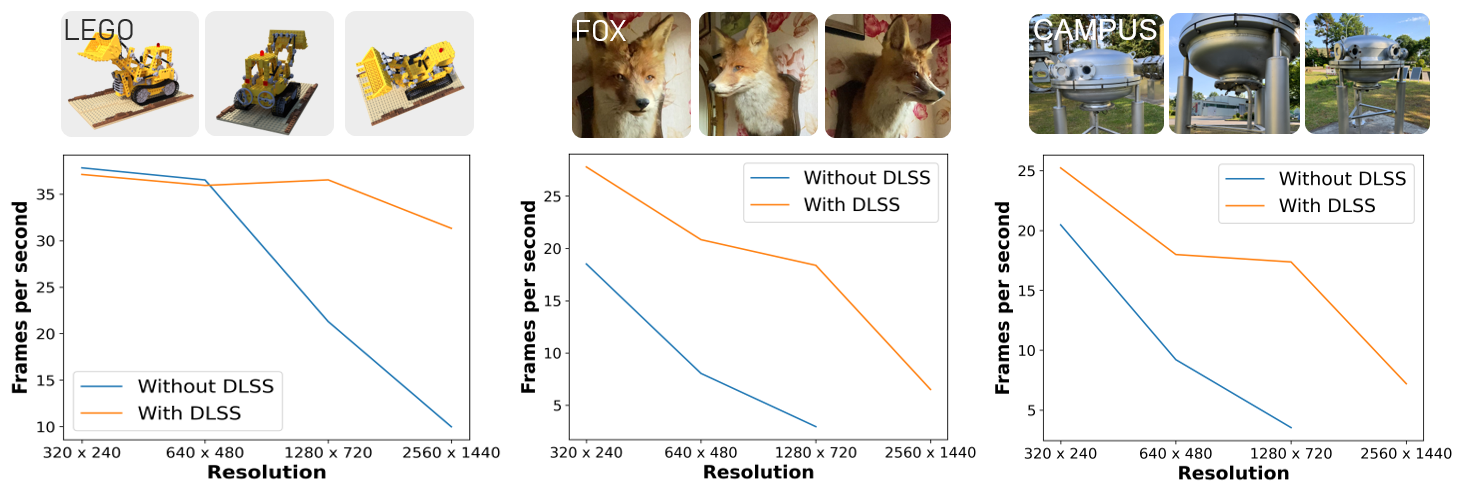}
    \caption{Illustration of the framerates for the three different datasets with different resolutions (cf. \autoref{sec:Benchmark Setup}) with and without DLSS support. For the datasets we choose: Lego (left), fox (middle) and campus (right).}
    \label{fig:Results}
\end{figure*}

\input{tables/table.tex}

\subsection{Benchmark Setup}
\label{sec:Benchmark Setup}
%
%
%
To evaluate the overall performance of our framework, we benchmarked the average, upper quantile of the fastest 25\%, and lower quantile of the slowest 25\% of frame timings. We employed three different scene complexities to evaluate our framework under varying viewing conditions. The scene with the least complexity is the Lego model of the original NeRF paper \cite{NERForiginal}, containing 100 images of $800\times 800$ pixel resolution, and an axis-aligned bounding box (AABB) size of 2. The medium-sized scene is the fox scene of instant-ngp \cite{instant-ngp} containing 50 images with a resolution of $1920\times 1080$ pixels and an AABB size of 4. The most complex scene, referred to as campus, is a custom dataset we collected, including multiple objects with dense information in the background. It contains 126 images with a resolution of $4032\times 3024$ pixels and an AABB size of 16. An example of the input images at three different angles is illustrated in Figure \ref{sec:Performance Benchmarking}.

We further evaluated how well our framework performs at four different resolution levels: \mbox{$320\times 240$}, \mbox{$640\times 480$}, \mbox{$1280\times 720$}, and \mbox{$2560\times 1440$}. 
In addition, we benchmarked against all scenes and resolutions with and without DLSS enabled.
To reduce bias, we ran each benchmark three times and restarted Unity after each benchmark to avoid internal caching.
For each run, we ran the application for at least one minute and only explored the areas with dense visual information to simulate a realistic usage of the framework.  
We chose an Oculus Quest 2 with a native rendering resolution of \mbox{$1920\times 1832$}, which we connected via a link cable to secure data streaming performance.

We utilized the Unity profiler\footnote{https://docs.Unity.com/Manual/Profiler.html} and Profiler Analyzer\footnote{https://docs.Unity.com/Packages/com.unity.performance.profile-analyzer@0.4/manual/index.html}, which can capture the timings of every frame and analyze them directly in Unity. 
All data was measured on a computer with a Quadro RTX 8000 GPU, an i7 4730K CPU, and 16GB of RAM.

\subsection{Benchmark Results}

\autoref{fig:Results} plots the frame rate versus render resolution per eye measured in fps for the three benchmark scenes. Each graph illustrates the trend of frame rate performance with and without DLSS support. Table \ref{tab:Results} enumerates each measurement result in terms of frame rate and frame timings with the latter measured in milliseconds (ms). 

The highest performance (averaged over all scenes) both in terms of frame timings and the framerate was measured for the lowest resolution of $320\times 240$ when DLSS was enabled. With increasing render resolution, frame rates decreased and frame timings increased. Moreover, rendering speed and frame rate became lower with higher scene complexity due to denser volume information for scene representation. As a result, the Lego scene with DLSS support had the best performance, providing a stable average frame rate above 30 fps, regardless of measured resolution. This allows real-time immersive NeRF rendering of the Lego scene at as high as $2560\times 1440$ per-eye resolution. Although with larger scenes, render time and frame rate could be limited by the current GPU capability, adding DLSS support significantly improved the performance such that our framework could achieve low-latency, high-resolution rendering with reasonable frame rates at 20 fps for the medium and larger scene as well. 


\section{Limitations}
\label{sec:Limitations}

While our framework opens up the possibility to render NeRF scenes in VR, the current implementation still underlies some limitations.  
First, direct interaction with individual objects within the NeRF scene is not possible yet, which could limit the interactivity of the VR experience. For example, this makes locomotion by teleportation challenging, as there is no collision data for specifying a teleportation way-point. While it is possible to generate a polygonal mesh from the volumetric density using methods such as the marching cube algorithm \cite{MarchingCubeAlgorithm}, the generated mesh currently has limited quality and thus could limit the accuracy of physics interactions. It remains a challenging research question how to create accurate physics collision and interactions within a NeRF scene.

The second main limitation of the current framework is that we observed a drop in frame rates below 30 fps for medium- and high-complexity scenes. Additional benchmarks would be necessary to evaluate whether this can be counteracted by either utilizing specialized hardware, like tensor processing units, or more powerful GPUs providing higher computational capabilities. 
Foveated rendering might also improve the performance as suggested by Deng et al. \cite{FoV-NeRF}.




\section{Conclusion and Future Work}
\label{sec:Conclusion}


In this paper, we presented an instant neural graphics primitives implementation for immersive NeRF in VR environments. A benchmark of scenes with different complexities and varying resolutions revealed frame rates up to 38 fps, making our approach feasible especially for rendering small-sized scenes.
In comparison to other techniques, such as photogrammetry, our framework enables rapid digitalization of real-world 3D environments from only unconstrained 2D images. Furthermore, using NeRF as a 3D scene representation can compress the scene data as the input images of the originally captured (potentially large-scale) environments are translated into network weights that can be stored with less memory usage.\\

There are multiple future extensions of our proposed system.
While our benchmark is covering objective system performance measurements, an empirical user evaluation could provide further insights into the perception of the created immersive NeRF scenes, for example, in terms of presence, motion sickness, perceived depth and sizes, spatial orientation, as well as induced motion sickness. 
In particular, a comparison with point-cloud or mesh-based techniques would be important to validate the suitability of immersive NeRF for application fields with varying requirements.

We also see the potential to fuse NeRF with other types of immersive VR experiences to achieve photo-realistic augmentation of the VE, for example, by merging NeRF rendering results with high-resolution stereoscopic VST. Other types of fusion, such as combining the NeRF rendering result with a geometric model that only exhibits an abstract scene representation, for example, a CAD model, could generate a virtual environment that has both photo-realistic textures and an accurate polygon mesh for physics interactions.

The current implementation of our approach based on Unity enables extensions for various HMDs and sensors, for example, foveated rendering, dynamic FoV, or user interaction.



\acknowledgments{This work was supported by DASHH (Data Science in Hamburg - HELMHOLTZ Graduate School for the Structure of Matter) with the Grant-No. HIDSS-0002, and the German Federal Ministry of Education and Research (BMBF).}

\bibliographystyle{abbrv-doi}

\bibliography{template}
\end{document}

%% file: tables/table.tex
\begin{table*}
\centering
\caption{Frame timings in milliseconds (left) and frame rate in frame per second  (right) for all measured datasets with \emph{DLSS enabled/disabled} (cf. \autoref{sec:Benchmark Setup}). All runs that did not produce measurable data are marked as DNF. Note that a lower frame time is preferred.}
\resizebox{\textwidth}{!}{%
\begin{tabular}{c|c|c|c|c|c|c|c|c|c|c|c|c}
\toprule
& \multicolumn{6}{c|}{Frame timings} & \multicolumn{6}{c}{Frame rates}\\
& \multicolumn{3}{c}{DLSS enabled} & \multicolumn{3}{c|}{DLSS disabled} & \multicolumn{3}{c}{DLSS enabled} & \multicolumn{3}{c}{DLSS disabled}\\
Resolution\textbackslash Scene & Lego & Fox & Campus & Lego & Fox & Campus & Lego & Fox & Campus & Lego & Fox & Campus\\
\midrule
320 x 240 & 26.94 & 35.98 & 39.65 & 26.44 & 54.03 & 48.85 & 37.11 & 27.79 & 25.22 & 37.82 & 18.51 & 20.47\\
640 x 480 & 27.83 & 48.02 & 55.6 & 27.38 & 124.34 & 108.59 & 35.93 & 20.83 & 17.99 & 36.52 &  8.04 &  9.21\\
1280 x 720 & 27.38 & 54.41 & 57.58 & 46.95 & 338.05 & 282.57 & 36.53 & 18.38 & 17.37 & 21.30 &  2.96 &  3.54\\
2560 x 1440 & 31.93 & 153.68 & 138.82 & 100.14 &  DNF & DNF & 31.32 &  6.51 &  7.20 &  9.99 &  DNF  &  DNF\\
\bottomrule
\end{tabular}%
}
\label{tab:Results}
\end{table*}